# Optimized Three Deep Learning Models Based-PSO Hyperparameters for Beijing PM2.5 Prediction

Andri Pranolo [a, b, 1, *], Yingchi Mao [a, 2], Aji Prasetya Wibawa [c, 3],
Agung Bella Putra Utama [c, 4], Felix Andika Dwiyanto [c, 5]

[a] *Department of Computer and Technology, College of Computer and Information, Hohai University*
*1 Xikang Road, Nanjing, Jiangsu 211100, China*

[b] *Department of Informatics, Faculty of Industrial Technology, Universitas Ahmad Dahlan*
*Jl. Prof. Dr. Soepomo, S.H., Janturan, Warungboto, Umbulharjo, Yogyakarta 55164, Indonesia*

[c] *Department of Electrical Engineering, Faculty of Engineering, Universitas Negeri Malang*
*Jl Semarang 5, Malang, East Java 65145, Indonesia*

[1] andri.pranolo@tif.uad.ac.id *; [2] maoyingchi@gmail.com; [3] aji.prasetya.ft@um.ac.id;
[4] agungbpu02@gmail.com; [5] felix@ascee.org
* corresponding author

## ARTICLE INFO

## ABSTRACT

Deep learning is a machine learning approach that produces excellent performance in various applications, including natural language processing, image identification, and forecasting. Deep learning network performance depends on the hyperparameter settings. This research attempts to optimize the deep learning architecture of Long short term memory (LSTM), Convolutional neural network (CNN), and Multilayer perceptron (MLP) for forecasting tasks using Particle swarm optimization (PSO), a swarm intelligence-based metaheuristic optimization methodology: Proposed M-1 (PSO-LSTM), M-2 (PSO-CNN), and M-3 (PSO-MLP). Beijing PM2.5 datasets was analyzed to measure the performance of the proposed models. PM2.5 as a target variable was affected by dew point, pressure, temperature, cumulated wind speed, hours of snow, and hours of rain. The deep learning network inputs consist of three different scenarios: daily, weekly, and monthly. The results show that the proposed M-1 with three hidden layers produces the best results of RMSE and MAPE compared to the proposed M-2, M-3, and all the baselines. A recommendation for air pollution management could be generated by using these optimized models.

## I. Introduction

In air quality monitoring systems, PM2.5 concentration is a crucial measure. As public awareness rises, analyzing and anticipating pollution levels is vital. Monitoring stations can only perform a small role in PM2.5 pollution control due to the nonlinear character of PM2.5 concentrations in both time and space. As a result, improving PM2.5 concentrations prediction accuracy is crucial for preventing and controlling air pollution. Several studies have been conducted using machine learning techniques, such as neural networks, applied to environmental science issues.

As a part of a neural network, deep learning is a technique that achieves high performance for various applications such as natural language processing, visual recognition, and forecasting has recently gained attention in the machine learning field. Machine learning models are characterized by large hyperparameter spaces and lengthy training times in their application. These properties, combined with the growth of parallel computing and the increasing demand for producing machine learning workloads. Therefore, developing mature hyperparameter optimization functionality for distributed computing environments is vital.

In most cases, machine learning provides more sensible advice than humans can. The design and training of neural networks, called alchemy, are tricky and unpredictable [1]. Therefore, hyperparameter tuning has been extensively studied to lower entry barriers for non-technical users.

Hyperparameter refers to parameters that cannot be changed during machine learning training. It can be involved in the model structure, such as the hidden layer and the activation function.

Two recent deep learning model development has made hyperparameter an increasingly important technique. The first is the scaling up of neural networks to achieve greater accuracy [2], and the second is the development of an intricate lightweight model to achieve greater accuracy with fewer data and parameters [3][4]. Furthermore, hyperparameter tuning plays an essential role in both cases. In its application, there are more hyperparameters to tune in a model with a complex structure than in a model with a well-defined structure. Several hypermeters for an LSTM model are necessary to improve performance, such as the number of hidden layers and neurons, dense layer, and weight initialization.

The first consideration of hyperparameter is the number of nodes and hidden layers. Hidden layers are the layers between the input and output layers. No specific number of hidden layers in its application should be used. Therefore, it depends on each problem to use a trial-and-error tuning approach. One hidden layer will suffice for most simple problems, and for more complex ones, two layers are recommended. Even though many nodes within a layer can improve accuracy, fewer nodes may result in underfitting [5]. The next are some units in a dense layer, which is the most used layer and essentially layer where all neurons as input for each neuron in the prior densely connected layer, which can increase the accuracy, while 5–10 units or nodes per layer is an ideal starting point for dense layers. As a result, the final dense layer's shape is influenced by the number of neurons/units specified [6].

Then, a dropout layer should be present between each LSTM layer, such as a layer that reduces the network's sensitivity to specific weights of individual neurons. The dropout layer can be used with an input layer. However, it cannot be used with the output layer since it can make the model and calculation errors. The dropout can alleviate the risk of overfitting when adding complexity by increasing the number of nodes in dense layers or adding more dense layers, resulting in poor validation accuracy [7]. In other cases, weight initialization can be a hyperparameter that should be considered. Ideally, the weight initialization schemes should differ depending on the activation function. However, weight values are chosen using a uniform distribution. Initially, it is impossible to set all weights to 0.0 because the optimization algorithm highlights the asymmetry in the error gradient. Different weights can lead to different starting points for the optimization process, leading to different final sets with different performance characteristics [8]. Stochastic optimization assumes that weights will be randomly assigned to small numbers at the start of the search.

As long as there is no weight update, weight decay can be included in the weight update rule. The weights are multiplied with slightly less than one factor to limit the weight growth. For references, the initial value of 0.97 should be sufficient. Moreover, the output of a node is defined by its activation functions, either ON or OFF. Using these functions, deep learning models can learn nonlinear prediction boundaries. Although it is technically possible to include activation functions in the dense layers, it is preferable to separate them into separate layers so that it could be reduced density layer output. The activation layer's choice depends on the application, but the most popular activation function is the rectifier [8].

The next hyperparameter is a learning rate. By using this hyperparameter, the network can update its parameters more quickly. To speed up the learning process, it is possible that increasing the learning rate will cause the model to diverge or even fail to converge. Learning will take longer, but the model will smoothly converge [9]. Alternatively, this hyperparameter is used in the training phase, with values between 0.0 and 0.1. Then, this hyperparameter specifies the number of epochs (integer) until the validation accuracy decreases even though training accuracy increases, thus risking overfitting. An ideal move is to use the early stopping method to specify the epochs number and stop training when the performance of the approach on the trained dataset drops below a pre-set threshold. The last consideration of hyperparameter tuning is batch size. This hyperparameter specifies the number of samples before updating internal model parameters. A more extensive sample size produces more significant gradient steps than smaller ones. The initial batch size is 32. However, it can adjust with multiples of 32, such as 64, 128, and 256, to determine which is better [8].

The research reveals that the PSO optimized deep learning models (LSTM, CNN, and MLP) for Beijing PM2.5 multivariate time series prediction acquire a minimum error and improve its accuracy. The seven optimizer hyperparameters are the optimizer, type of activation function, loss function, number of batch sizes, hidden units, neurons, and epochs.

The contribution of the research are:

1) To improve the accuracy of the multivariate time-series forecasting analysis applied to Beijing PM2.5 dataset using the proposed model M1 (PSO-LSTM), M2 (PSO-CNN), and M3 (PSO-MLP).
2) To generate the computer-based forecasting model that could as a recommendation for governmental regulations such as pollution prevention, Clean Air Technology Center, and transportation-emissions reduction.

The research may present the alternative use of PSO as a tuning hyperparameter on deep learning instead of using it as a feature selection. The automatic tuning process may reduce the computational time due to the random parameter selection. Finally, this paper determines the best optimized deep learning approaches to predict Beijing PM2.5 concentrations.

## II. Method

The proposed hyperparameter tuning of deep learning for forecasting is shown in Figure 1. As shown, the selected dataset will be preprocessed using normalization. The use of the PSO carries out the hyperparameter selection. The best-selected hyperparameter values will be used in the forecasting. Then, the forecasting process will take place by a deep learning method, namely LSTM, CNN, and MLP. In the end, the proposed models and the baseline performance were tested using MAPE and RMSE.

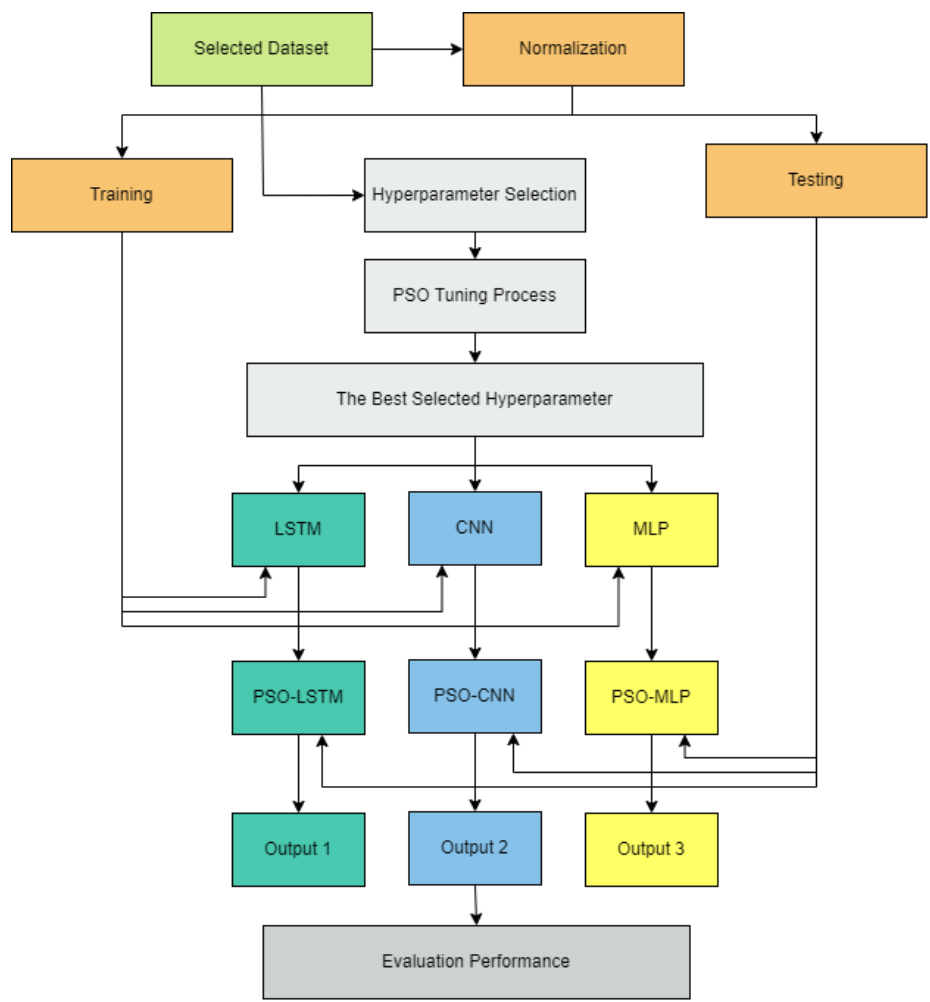

Fig. 1. The proposed hyperparameter tuning of deep learning for forecasting

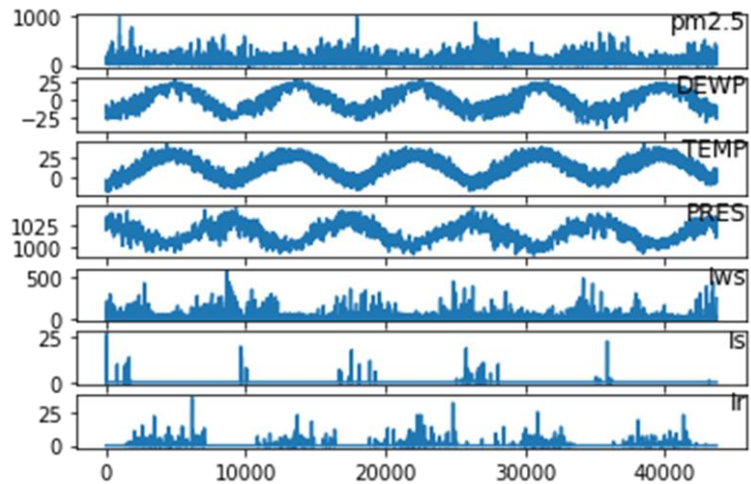

Fig. 2. Visualize the dataset of Beijing PM2.5

*A. Dataset*

In this study, an evaluation of the hyperparameter setting of the LSTM method based on the PSO Dataset using PM2.5 Beijing was carried out, which was obtained from the UCI machine learning repository [10]. This dataset represents the weather conditions, and pollution levels reported hourly by the US. in Beijing, China, from 2010 to 2014, with 43.825 instances removed and 2.068 data row values missing in data preprocessing. Pre-processing is the initial process of datasets to improve data quality and selection to obtain high-performance results.

Preprocessing data used are feature selection and data normalization. The feature selection process selects the attributes to be used by following a similar study conducted by Zhang [11] using seven attribute data features which include PM2.5 concentration (pm2.5), dew point (DEWP), temperature (TEMP), pressure (PRES), accumulated wind speed (lr), hourly snow accumulation (l), and hourly rain accumulation (lr) as shown in Figure 2. Normalization is a technique for reducing errors by converting the real number to a value range of 0 to 1. The min-max scaling approach is used for normalization [12]. Equation (1) presents the normalization min-max.

$$X_{i(norm)} = \frac{X_i - X_{min}}{X_{max} - X_{min}} \tag{1}$$

$X_{i(norm)}$ is a normalization result, $X_i$ represents the data to be normalized while $X_{min}$ and $X_{max}$ is the values of minimum and maximum of entire data. In this study, from the dataset, there were three scenarios used as testing data. They are monthly, weekly, and daily.

*B. Hyperparameter Optimized using PSO*

Developing an efficient machine learning model is a complex process that requires selecting a suitable algorithm and modifying the model's hyperparameters [13]. The primary goal of hyperparameter optimization is to simplify the selection of parameters to get the optimal results of the process and enable users to implement efficient machine learning models to solve practical issues [14].

The process of hyperparameter optimization predicts the best machine learning (ML) architecture [15]. It decreases the amount of human work necessary, enhances machine learning models' performance, and increases models' reproducibility. Particle swarm optimization (PSO) is a swarm optimization model that could use to select hyperparameters and is used in this research as an integrated approach with the other baseline deep learning models.

PSO is a family of evolutionary algorithms frequently used to solve optimization problems and has been effectively applied as parameter optimization techniques [16]. PSO takes its inspiration from biological populations that exhibit individual and social behavior. PSO works by allowing a swarm of particles to navigate semi-random search space. Through integrated information sharing between individual particles in a group, PSO algorithms determine the optimal solution.

In PSO, a swarm $S$ consists of a group of $n$ particles [17] as in (2), and a vector is used to represent each particle $S_i$, as seen in (3).

$$S = (S_1, S_2, \ldots, S_n) \qquad (2)$$

$$S_i = <\vec{x_i}, \vec{v_i}, \vec{p_i}> \qquad (3)$$

where $\vec{x_i}$ denotes the current position, $\vec{v_i}$ denotes the current velocity, and $\vec{p_i}$ denotes the swarm's best-known position. After initializing each particle's position and velocity, the current position and records are analyzed with their performance score. The following iteration modifies the velocity $\vec{v_i}$ of each particle following the current global optimal position $\vec{p}$ and the prior position $\vec{p_i}$, as in (4).

$$\vec{v_i} := \vec{v_i} + U(0, \varphi_2)(\vec{p} - \vec{x_i}) + U(0, \varphi_1)(\vec{p_i} - \vec{x_i}) \qquad (4)$$

where $U(0, \varphi)$ denotes distributions of continuous uniform based on the $\varphi_1$ and $\varphi_2$ acceleration constants. Equation (5) represents that the particles move following their new velocity vectors.

$$\vec{x_i} := \vec{x_i} + \vec{v_i} \qquad (5)$$

The technique outlined above is performed until convergence or termination constraints are met. The PSO algorithm has a computational complexity of $O(n \log n)$ [18]. Additionally, this approach can be parallelized to increase model efficiency because PSO particles act independently and share information only after each iteration.

PSO's primary restriction requires adequate population initialization. It may reach a local rather than global optimum in discrete hyperparameters [19]. In carrying out the appropriate population initialization, using population initialization techniques or utilizing the developer's experience is necessary. Numerous population initialization strategies, such as the opposition-based optimization algorithm [20] and the space transformation search approach [21] have been developed to increase the performance of evolutionary algorithms. Thus, execution time and resource optimization can be increased by performing an extra population initialization strategy

Through hyperparameter selection, PSO can improve good values of Deep learning (DL) models. DL is based on artificial neural network theory (ANN). Multilayer perceptrons (MLP), convolutional neural networks (CNNs), recurrent neural networks (RNN), Deep neural networks (DNN), and long short-term memory (LSTMs) are modified from the standard of ANN for deep learning designs [22]. The hyperparameters in the DL that PSO can optimize for selecting hyperparameters include the optimizer, activation function, loss function, batch size, number of neurons, and epochs.

Hyperparameters tuning with PSO can be done by calling the optimal configuration 'particle swarm' in the opportunity function in the TensorFlow Keras package. The used PSO parameters consist of 10 particles in the swarm, 5 generations (iterations), velocity minimum 0, velocity maximum 1, $pBest$ 1.5, $gBest$ 2.0, and 10 permitted function evaluations. The hyperparameters optimized by tuning PSO and retested using the Deep Learning method can be seen in Table 1 and applied a dropout value of 0.2. The parameters that are tuned are parameters that are shared by all deep learning methods in general.

Table 1. Deep learning method hyperparameter space

| No. | Hyperparameters | Search Space | Type |
| --- | --- | --- | --- |
| 1. | Hidden layers (HL) | [2,10] | Continuous |
| 2. | Neurons | [1,100] | Continuous |
| 3. | Activation function | Linear, Sigmoid, ReLU | Discrete with step=1 |
| 4. | Loss function | MSE, MAE | Discrete with step=1 |
| 5. | Optimizer | Adam, RMSprop | Discrete with step=1 |
| 6. | Batch size | [32, 64, 128] | Discrete with step=1 |
| 7. | Epoch | [5,100] | Continuous |

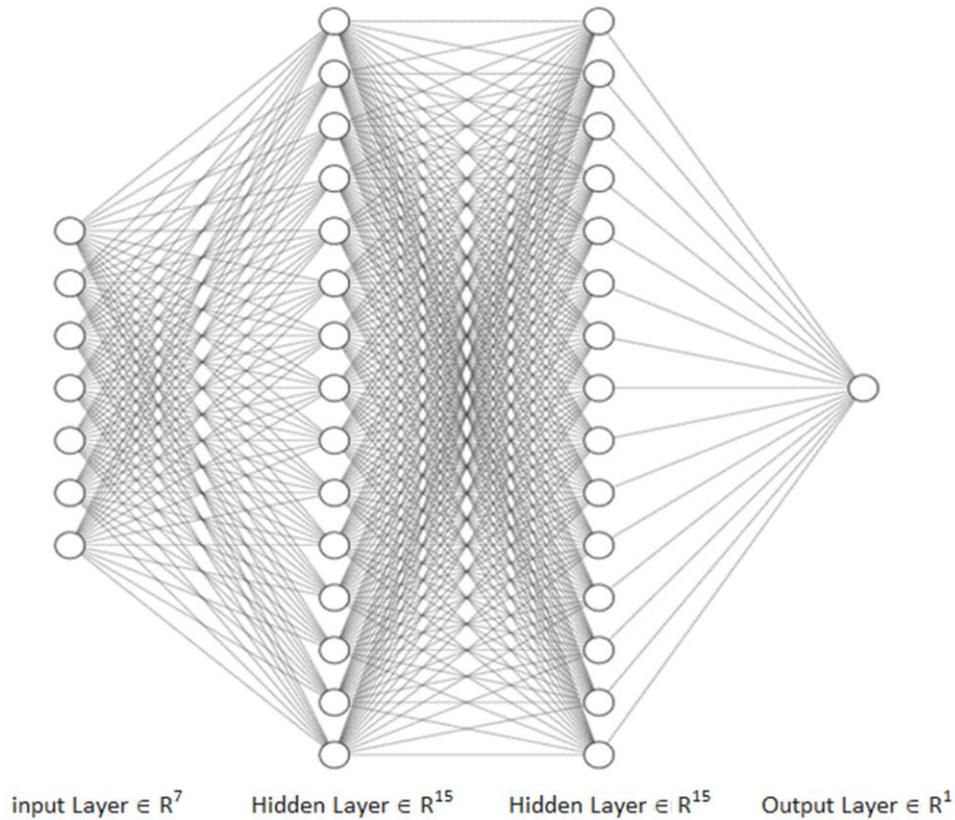

Fig. 3. MLP architecture

*C. Multilayer Perceptron (MLP)*

The forecasting method often used in research is MLP [23]. MLP belongs to the feedforward network. The characteristics possessed by MLP are advantages in determining the value of weights that are better than other methods, MLP can be used without prior knowledge, and the algorithm can be implemented quickly and can solve linear and nonlinear problems [24]. MLP characteristics make the forecasting value better. MLP in forecasting is used for time series [25] and stock prices [26][27].

As illustrated in Figure 3, the MLP model architecture consists of three layers of nodes: an input layer, a hidden layer, and an output layer. Each layer is connected to the network architecture nodes. The nodes in the input layer are connected to nodes in the hidden layer, and the hidden layer's nodes are directly connected to nodes in the output layer's node. The elements of a multilayer perceptron consist of network architecture, learning algorithms, and activation functions [28].

Activation function for an $i^{th}$ in a hidden neuron could be defined as in (6).

$$h_i = f(u_i) = f\left(\sum_{k=0}^{K} w_{ki} x_k\right) \tag{6}$$

where $h_i$ is hidden neuron of $i^{th}$, $f(u_i)$ denotes a link function that adds non-linearity to the relationship between the input and hidden layers, $w_{ki}$ denotes $(k,i)^{th}$ weight as input in a weight (K X N) matrix, $x_k$ is $K^{th}$ represents an input value. $y_j$ is $j^{th}$ output values as in (7).

$$y_j = f(u'_j) = f\left(\sum_{i=0}^{N} w'_{ij} h_i\right) \tag{7}$$

*D. Long Short-Term Memory (LSTM)*

Long Short-Term Memory (LSTM) is developed from the recurrent neural network (RNN) that could implement to solve the problem of accuracy in time-series data prediction. LSTM can overcome long-term dependencies on its inputs [29]. LSTM creates RNN architectures capable of resolving learning challenges associated with information linkage. In an RNN, the old memory becomes increasingly ineffective as the new memory overwrites it [30]. However, RNNs suffer from vanishing and bursting gradients, which occur when the range of values across layers in architecture changes. The LSTM was developed and designed to address the issue of RNN gradient disappearing while faced with vanishing and bursting gradients [31]. Time series forecasting using LSTM can be used for time-series predictions [32], both short-term loads [33] or long-term [34], weather predictions [35], price movements [36][37][38][39].

The LSTM uses memory cells and gate units to manage memory at each input, with an architecture similar to the RNN. In LSTM, the hidden layer comprises memory cells with three gates: input, forget, and output, as illustrated in Figure 4. The input gate specifies the amount of data stored in the cell state and keeps the cell from holding extraneous data. Forget gate functions limit the time a value remains in a memory cell. The output gate determines the amount of data or value stored in a memory cell and calculates the output.

On the LSTM, the gate is a unique network structure with an input vector and output intervals of 0 and 1. No information is permitted to flow when output is set to 0. In contrast, all information is permitted to pass when set to 1 [40]. If the input vector $x = (x_1, x_2, \ldots, x_{t-1}, x_t)$ and output vector $(S_1, S_2, \ldots, S_{t-1}, S_t)$ are defined, then *gates* could be formulated as in (8).

$$g(x) = \sigma(Wx + b) \tag{8}$$

Sigmoid $\sigma(x) = 1/(1 + e^{-x})$; where $W$ denotes the weights and $b$ denotes the bias vector. The cell state represents the current condition of the cell as being determined as (9).

$$C_t = f_t \cdot C_{t-1} + i_t \cdot \tanh(W_c \cdot [S_{t-1}, X_t] + \text{bc}) \tag{9}$$

$W_c$ denotes the cell state matrix's weight, $b_c$ denotes the cell state's bias vector as the input gate, and $f_t$ is the forget gate used to assist the network in forgetting input information and repeating memory cells. The input and forget gates can be computed using the formulas (10) and (11).

$$i_t = \sigma(W_i \cdot [S_{t-1}, X_t] + \text{bi}) \tag{10}$$

$$f_t = \sigma(W_f \cdot [S_{t-1}, X_t] + \text{bf}) \tag{11}$$

$W_i$ and $W_f$ denote the weights of the input and forget gates, respectively, while bi denotes the bias vectors of the input-gate, and bf denotes the forget-gate bias vectors. The output-gate of the LSTM regulates the amount of information processed into the output from the latest cell state. The output

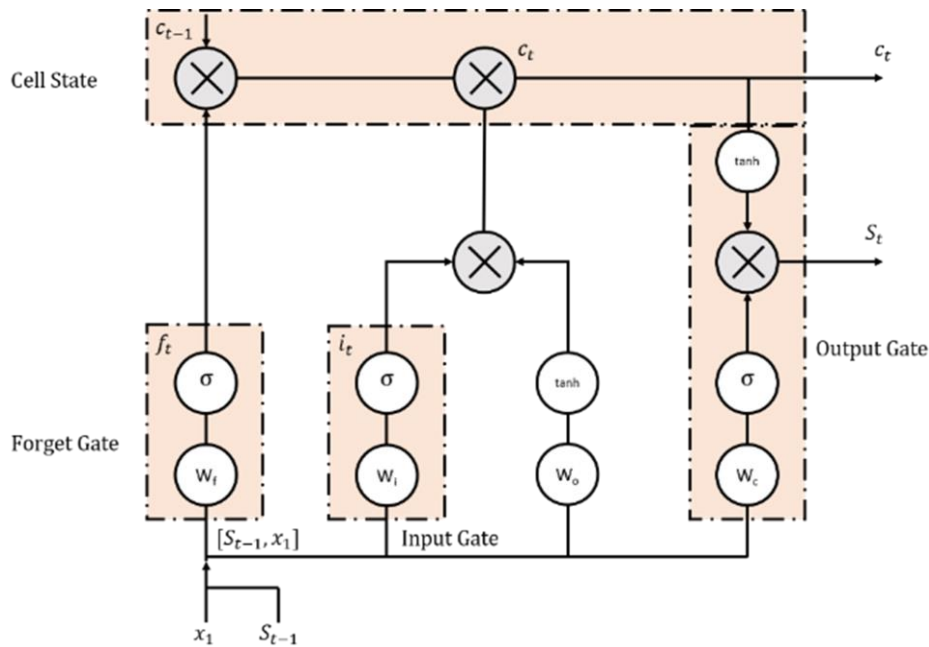

Fig. 4. Memory cells LSTM

can be estimated using the formula in (12).

$$O_t = \sigma(W_o \cdot [S_{t-1}, X_t] + bo) \qquad (12)$$

$W_o$ denotes output gate matrix weight, and *bo* is the gate output bias vector. The LSTM process's ultimate output is computed as in (13).

$$s_t = o_t \cdot \tanh(c_t) \qquad (13)$$

Then the output will be used for forecasting the following time chosen.

### E. Convolutional Neural Network (CNN)

CNN is part of the DL approach, which is included in the sub-field of ML, which applies the basic concepts of the ANN algorithm with more layers [41]. CNN is a feedforward network because information flow occurs in one direction only, from their inputs to their outputs CNN was applied and extremely popular in image classification research. Therefore, it could be implemented for 1-dimensional (1D) problems, such as forecasting the following values in a time series dataset [42]. The model used is a 1D CNN with architecture, as in Figure 5.

Many types of CNN models can be used for each problem in predicting data time series. The model consists of univariate, multivariate, multi-step, and multivariate multi-step [43]. CNN in forecasting data is time series often used to estimate stock prices [44][45], gold prices [46][47][48], health [49][50][51], time series [52][53][54], solar cells and weather forecasts [55].

### F. Evaluation

The mean absolute percentage error (MAPE) as error evaluation metrics and the root mean square error (RMSE) [56] was used to evaluate and compare the implemented methods' performances. MAPE shows errors that can represent accuracy. At the same time, RMSE detect irregularities or outliers in the designed projection system. The formulas are given as in (14) and (15).

From the calculation of the MAPE and RMSE value, it will be known which model has the best performance in forecasting. The smaller MAPE and RMSE values produced, the better the forecasting results, so the method was better [57].

$$MAPE = \sum_{t=1}^{n} \frac{|e_t|}{n.X_t} \times 100 \qquad (14)$$

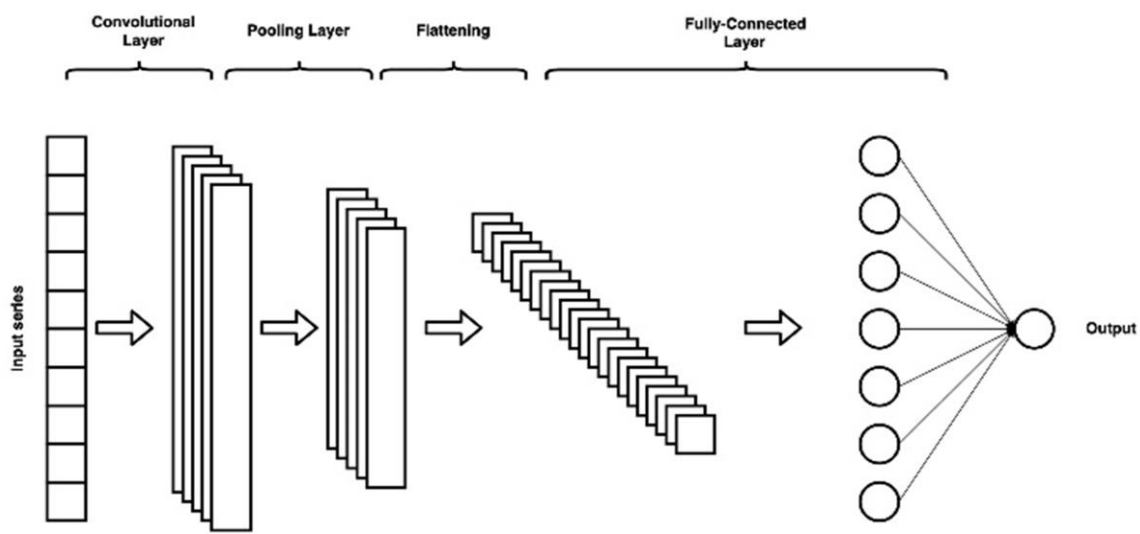

Fig. 5. 1D CNN architecture

$$RMSE = \sqrt{\sum_{t=1}^{n} \frac{(e_t)^2}{n}} \tag{15}$$

### III. Results and Discussion

The original deep learning (LSTM, CNN, and MLP) architecture are 7 input layers, 2 to 10 hidden layers (HL), and 1 output layer with the same setting parameter values. The parameters are 32 neurons, dropout 0.2, MSE for loss function, Adam Optimizer, 100 epoch, and 72 batch size. Unlike LSTM and MLP, CNN used the parameters in the fully connected layer. The specific CNN architecture setting uses 1D convolution layer with 2 kernel sizes, 64 filters, ReLU for activation function, pooling layer with MaxPooling1D type, size 1, and drop out 0.2. Then there was 1 flattened layer and a fully connected layer. Based on the current tuning results, test the PSO tuning results using the Deep Learning method with the result settings as shown in Table 2.

PSO hyperparameter tuning was integrated with various deep learning models (LSTM, CNN, and MLP) to produce new models of Proposed Model M-1 (PSO-LSTM), M-2 (PSO-CNN), and M-

Table 2. PSO hyperparameter search results deep learning method

| No. | Hyperparameters | Proposed M-1 | Proposed M-2 | Proposed M3 |
|---|---|---|---|---|
| 1. | Hidden layers (HL) | 3 | 4 | 3 |
| 2. | Neurons | 24 | 41 | 61 |
| 3. | Activation function | Sigmoid | ReLU | Linear |
| 4. | Loss function | MSE | MAE | MSE |
| 5. | Optimizer | Adam | RMSprop | RMSprop |
| 6. | Batch size | 32 | 32 | 64 |
| 7. | Epoch | 46 | 60 | 68 |

Table 3. MAPE forecasting results

| Model | MAPE | | | | | | | | |
|---|---|---|---|---|---|---|---|---|---|
| | HL-2 | HL-3 | HL-4 | HL-5 | HL-6 | HL7 | HL-8 | HL-9 | HL-10 |
| **Monthly** | | | | | | | | | |
| LSTM | 9.1216 | **8.8909** | 9.1385 | 9.1935 | 9.2448 | 9.2544 | 9.2612 | 9.2711 | 9.3865 |
| CNN | 8.6255 | 8.6195 | **8.5849** | 8.9762 | 9.1778 | 10.3037 | 10.7662 | 10.8264 | 11.1073 |
| MLP | 9.3308 | **9.2286** | 9.5347 | 9.6395 | 10.6010 | 10.6280 | 10.6702 | 10.6008 | 10.6035 |
| Proposed M-1* | - | **8.4576** | - | - | - | - | - | - | - |
| Proposed M-2** | - | - | **8.5281** | - | - | - | - | - | - |
| Proposed M-3* | - | **9.0930** | - | - | - | - | - | - | - |
| **Weekly** | | | | | | | | | |
| LSTM | 8.9777 | **8.8327** | 10.1041 | 10.2722 | 10.3538 | 11.5553 | 11.5812 | 11.5852 | 11.5940 |
| CNN | 9.8238 | 8.9021 | **8.8092** | 8.9096 | 9.1951 | 10.2191 | 11.7623 | 12.7759 | 13.3261 |
| MLP | 9.9057 | **9.7078** | 10.0382 | 11.6556 | 11.6180 | 11.6290 | 11.6234 | 11.6228 | 11.6118 |
| Proposed M-1* | - | **8.6379** | - | - | - | - | - | - | - |
| Proposed M-2** | - | - | **8.6987** | - | - | - | - | - | - |
| Proposed M-3* | - | **9.2903** | - | - | - | - | - | - | - |
| **Daily** | | | | | | | | | |
| LSTM | 5.5329 | **5.5306** | 5.5343 | 5.5351 | 7.7324 | 8.7756 | 9.0327 | 10.1076 | 10.4688 |
| CNN | 6.7490 | 6.9270 | **6.4845** | 6.8979 | 6.9833 | 6.8275 | 6.8986 | 6.8067 | 8.9088 |
| MLP | 6.4448 | **6.2857** | 7.4463 | 8.5707 | 8.5792 | 8.5765 | 8.5684 | 8.5739 | 8.5703 |
| Proposed M-1* | - | **5.4676** | - | - | - | - | - | - | - |
| Proposed M-2** | - | - | **6.3742** | - | - | - | - | - | - |
| Proposed M-3* | - | **6.0990** | - | - | - | - | - | - | - |

*the best selection parameter was hidden layer 3 (HL-3)*
**the best selection parameter was hidden layer 4 (HL-4)*

62         *A. Pranolo et al. / Knowledge Engineering and Data Science 2022, 5 (1): 53-66*Table 4. RMSE forecasting results

| Model | RMSE | | | | | | | | |
|---|---|---|---|---|---|---|---|---|---|
| | HL-2 | HL-3 | HL-4 | HL-5 | HL-6 | HL7 | HL-8 | HL-9 | HL-10 |
| **Monthly** | | | | | | | | | |
| LSTM | 0.0260 | **0.0257** | 0.0263 | 0.0265 | 0.0270 | 0.0952 | 0.0952 | 0.0952 | 0.0953 |
| CNN | 0.0362 | 0.0357 | **0.0351** | 0.0369 | 0.0429 | 0.0636 | 0.0668 | 0.0764 | 0.0773 |
| MLP | 0.0263 | **0.0262** | 0.0265 | 0.0266 | 0.0945 | 0.0944 | 0.0944 | 0.0945 | 0.0945 |
| Proposed M-1* | - | **0.0250** | - | - | - | - | - | - | - |
| Proposed M-2** | - | - | **0.0346** | - | - | - | - | - | - |
| Proposed M-3* | - | **0.0259** | - | - | - | - | - | - | - |
| **Weekly** | | | | | | | | | |
| LSTM | 0.0299 | **0.0297** | 0.0302 | 0.0303 | 0.0311 | 0.1182 | 0.1183 | 0.1183 | 0.1183 |
| CNN | 0.0523 | 0.0437 | **0.0412** | 0.0475 | 0.0497 | 0.0556 | 0.0927 | 0.1019 | 0.1092 |
| MLP | 0.0304 | **0.0302** | 0.0310 | 0.1185 | 0.1184 | 0.1184 | 0.1184 | 0.1184 | 0.1184 |
| Proposed M-1* | - | **0.0232** | - | - | - | - | - | - | - |
| Proposed M-2** | - | - | **0.0362** | - | - | - | - | - | - |
| Proposed M-3* | - | **0.0301** | - | - | - | - | - | - | - |
| **Daily** | | | | | | | | | |
| LSTM | 0.0041 | **0.0039** | 0.0043 | 0.0049 | 0.0091 | 0.0844 | 0.0845 | 0.0848 | 0.0852 |
| CNN | 0.0192 | 0.0172 | **0.0101** | 0.0188 | 0.0157 | 0.0178 | 0.0168 | 0.0181 | 0.0241 |
| MLP | 0.0056 | **0.0049** | 0.0109 | 0.0785 | 0.0772 | 0.0776 | 0.0788 | 0.0780 | 0.0786 |
| Proposed M-1* | - | **0.0023** | - | - | - | - | - | - | - |
| Proposed M-2** | - | - | **0.0031** | - | - | - | - | - | - |
| Proposed M-3* | - | **0.0031** | - | - | - | - | - | - | - |

*\* the best selection parameter was hidden layer 3 (HL-3)*
*\*\* the best selection parameter was hidden layer 4 (HL-4)*

3 (PSO-MLP). MAPE and RMSE measured the performances of the proposed model and its comparison with the baselines, as shown in Table 3 and Table 4, respectively.

In general, all proposed models have better accuracy performance for all monthly, weekly, and daily scenarios, is indicated by the minimum MAPE (Table 3) and RMSE (Table 4) values obtained by the three proposed models compared to the other models. More specifically, in the monthly scenario, Proposed M-1 has the best performance of the three proposed models, followed by M-2, and M-3, with MAPE values of 8.4576, 8.5281, and 9.0930, respectively. In addition, the RMSE value also shows the same order of performance for the three proposed models, namely 0.0250, 0.0346, and 0.0259, respectively. The same thing happened in weekly and daily scenarios. However, if it was sorted based on the scenarios, the accuracy of the three proposed models with the best performance was shown in the daily scenario, followed by weekly and monthly. The increasing amount of data and precise outliers or distance precision within values on the dataset has contributed to the proposed model performance.

Proposed M-1 (PSO-LSTM) can also reduce the yield value of RMSE and MAPE to be better than LSTM as a baseline model. The tuning results for M-2 (PSO-CNN) have better RMSE and MAPE values than CNN when the hidden layer is 4 (HL-4). As for the proposed M-3 (PSO-MLP), the use of HL-3 has a better evaluation value when compared to MLP.

From the overall results in Table 3 and Table 4, the best results can be visualized as shown in Figure 6 and Figure 7. Figure 6 demonstrates that, when compared to all other models, the proposed model has the best MAPE value in every scenario. In the Monthly scenario, proposed M-1 outperforms regular LSTM, CNN, and MLP with a MAPE of 8.4576. The weekly scenario's MAPE proposed M-1 has a superior MAPE than previous techniques, with a score of 8.6379. The MAPE generated by proposed M-1 in the daily scenario was 5.4676, which was likewise better and more effective than other techniques. Figure 7 shows that every proposed model has the best RMSE in every scenario. Compared to other models, the monthly scenario's RMSE of 0.025, which belongs to proposed M-1, has the best value. The best result for the RMSE proposed M-1 in the Weekly scenario is 0.0232, which is lower than the RMSE of other models. The RMSE value for the daily data ranges from 0.0023 (proposed M-1) to 0.0039 (LSTM), 0.0101 (CNN), and 0.0049 (MLP). Overall, it can be seen that the PSO hyperparameter tuning in this research case study can improve the baseline models' performance. The RMSE and MAPE evaluation values of the M-1 produce the best values in all scenarios (Monthly, Weekly, and Daily) compared to other proposed models and the baselines.

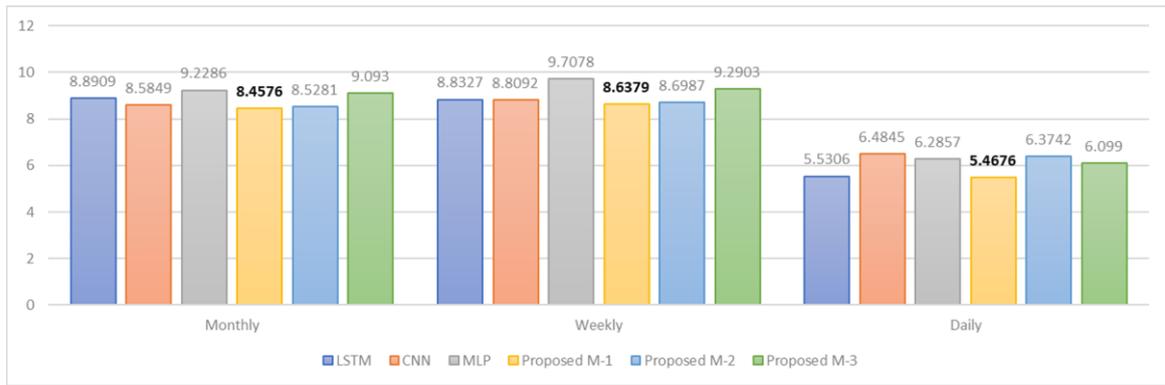

Fig. 6. Comparison of MAPE in all scenarios

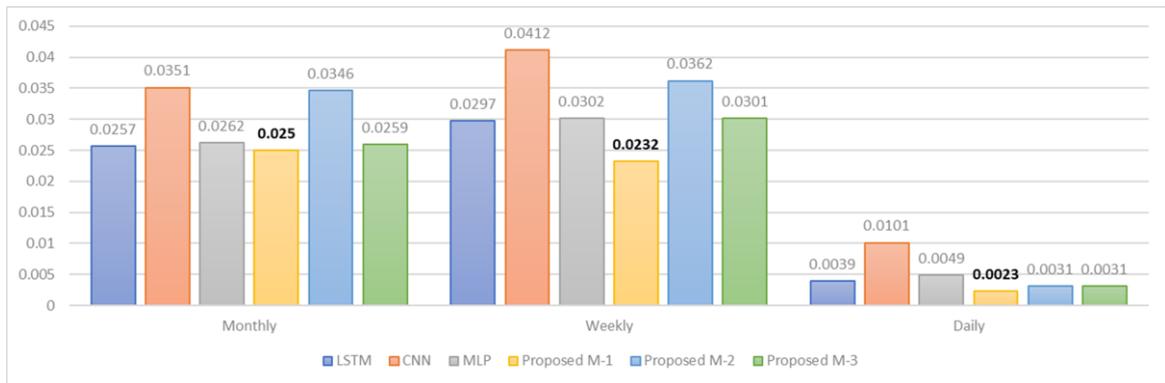

Fig. 7. Comparison of RMSE in all scenarios

The government may use this research finding to reference their regulations as a benefit of this research. The first regulation is pollution-prevention approaches aiming to minimize, remove, and avoid pollution. The government promotes the use of less hazardous raw resources or fuels, a less toxic industrial operation, and increased process efficiency. The second policy is to establish the Clean Air Technology Center, which will provide information on technologies for preventing and controlling air pollution, including mechanical collectors, fabric filtration, combustion systems, wet scrubbers, and biological degradation and their use, cost, and effectiveness. The third regulation reduces transportation-related emissions by requiring car emission controls and cleaner fuels. Finally, economic incentives for air pollution control agencies, such as emissions banking and trading, can be created.

## IV. Conclusion

This paper proposed improved deep learning approaches based on PSO hyperparameters tuning to select the best parameters. The experiment shows that all proposed models outperformed the baseline model. The best performance of Proposed M-1 (PSO-LSTM) outperformed other produced models, M-2 (PSO-CNN) and M-3 (PSO-MLP), and the baseline models, LSTM, CNN, and MLP. Governmental regulations such as pollution prevention, Clean Air Technology Center, and transportation-emissions reduction could be generated based on this promising finding. The proposed model in this study has good performance, which only applies to the dataset used.

Therefore, future research will use various datasets to produce a generally applicable model to all time-series datasets.

## Acknowledgment

The authors are grateful for the support provided by the Chinese Government Scholarship (CGS), which has contributed funding to conduct this research through the CGS Scholarship. In addition, appreciate Hohai University, Universitas Ahmad Dahlan, and Universitas Negeri Malang, which have contributed to supporting laboratory facilities.

## Declarations

*Author contribution*

All authors contributed equally as the main contributor of this paper. All authors read and approved the final paper.

*Funding statement*

This work is supported by the Chinese Government Scholarship (CGS) that received by the corresponding author with CSC Number 2018GBJ006341 and by Universitas Ahmad Dahlan under grant number PD-226/SP3/LPPM-UAD/VII/2022.

*Conflict of interest*

The authors declare no known conflict of financial interest or personal relationships that could have appeared to influence the work reported in this paper.

*Additional information*

Reprints and permission information are available at http://journal2.um.ac.id/index.php/keds.

Publisher's Note: Department of Electrical Engineering - Universitas Negeri Malang remains neutral with regard to jurisdictional claims and institutional affiliations.

## References

[1] T. Yu and H. Zhu, "Hyper-Parameter Optimization: A Review of Algorithms and Applications," *arXiv Prepr. arXiv2003.05689*, Mar. 2020.

[2] M. Tan and Q. V. Le, "EfficientNet: Rethinking Model Scaling for Convolutional Neural Networks," *arXiv Prepr.*, May 2019.

[3] N. Ma, X. Zhang, H. T. Zheng, and J. Sun, "Shufflenet v2: Practical guidelines for efficient cnn architecture design," in *Proceedings of the European conference on computer vision (ECCV)*, 2018, pp. 116–131.

[4] M. Sandler, A. Howard, M. Zhu, A. Zhmoginov, and L. C. Chen, "Mobilenetv2: Inverted residuals and linear bottlenecks," in *Proceedings of the IEEE conference on computer vision and pattern recognition*, 2018, pp. 4510–4520.

[5] X. Zhang, X. Chen, L. Yao, C. Ge, and M. Dong, "Deep Neural Network Hyperparameter Optimization with Orthogonal Array Tuning," in *Neural Information Processing*, T. Gedeon, K. Wong, and M. Lee, Eds. Springer, 2019, pp. 287–295.

[6] N. Gorgolis, I. Hatzilygeroudis, Z. Istenes, and L. n G. Gyenne, "Hyperparameter Optimization of LSTM Network Models through Genetic Algorithm," in *2019 10th International Conference on Information, Intelligence, Systems and Applications (IISA)*, Jul. 2019, pp. 1–4.

[7] G. E. Hinton, N. Srivastava, A. Krizhevsky, I. Sutskever, and R. R. Salakhutdinov, "Improving neural networks by preventing co-adaptation of feature detectors," *arXiv Prepr. arXiv1207.0580*, Jul. 2012.

[8] A. Farzad, H. Mashayekhi, and H. Hassanpour, "A comparative performance analysis of different activation functions in LSTM networks for classification," *Neural Comput. Appl.*, vol. 31, no. 7, pp. 2507–2521, Jul. 2019.

[9] M. D. Zeiler, "ADADELTA: An Adaptive Learning Rate Method," *arXiv Prepr. arXiv1212.5701*, Dec. 2012.

[10] X. Liang *et al.*, "Assessing Beijing's PM2.5 pollution: Severity, weather impact, APEC and winter heating," *Proc. R. Soc. A Math. Phys. Eng. Sci.*, vol. 471, no. 2182, 2015.

[11] M. Zhang, D. Wu, and R. Xue, "Hourly prediction of PM2.5 concentration in Beijing based on Bi-LSTM neural network," *Multimed. Tools Appl.*, vol. 80, no. 16, pp. 24455–24468, 2021.

[12] S. E. Buttrey, " Data Mining Algorithms Explained Using R ," *J. Stat. Softw.*, vol. 66, no. Book Review 2, 2015.

[13] R. Elshawi, M. Maher, and S. Sakr, "Automated Machine Learning: State-of-The-Art and Open Challenges," Jun. 2019.

[14] L. Yang and A. Shami, "On hyperparameter optimization of machine learning algorithms: Theory and practice," *Neurocomputing*, vol. 415, pp. 295–316, Nov. 2020.

[15] F. Hutter, L. Kotthoff, and J. Vanschoren, *Automated Machine Learning*. Cham: Springer International Publishing, 2019.

[16] N. Xue, I. Triguero, G. P. Figueredo, and D. Landa-Silva, "Evolving Deep CNN-LSTMs for Inventory Time Series Prediction," *2019 IEEE Congr. Evol. Comput. CEC 2019 - Proc.*, pp. 1517–1524, 2019.

[17] M.-A. Zöller and M. F. Huber, "Benchmark and Survey of Automated Machine Learning Frameworks," Apr. 2019.

[18] X.-H. Yan, F.-Z. He, and Y.-L. Chen, "A Novel Hardware/Software Partitioning Method Based on Position Disturbed Particle Swarm Optimization with Invasive Weed Optimization," *J. Comput. Sci. Technol.*, vol. 32, no. 2, pp. 340–355, Mar. 2017.

[19] M.-Y. Cheng, K.-Y. Huang, and M. Hutomo, "Multiobjective Dynamic-Guiding PSO for Optimizing Work Shift Schedules," *J. Constr. Eng. Manag.*, vol. 144, no. 9, p. 04018089, Sep. 2018.

[20] S. Rahnamayan, H. R. Tizhoosh, and M. M. A. Salama, "A novel population initialization method for accelerating evolutionary algorithms," *Comput. Math. with Appl.*, vol. 53, no. 10, pp. 1605–1614, May 2007.

[21] H. Wang, Z. Wu, J. Wang, X. Dong, S. Yu, and C. Chen, "A New Population Initialization Method Based on Space Transformation Search," in *2009 Fifth International Conference on Natural Computation*, 2009, pp. 332–336.

[22] M. Hiransha, E. A. Gopalakrishnan, V. K. Menon, and K. P. Soman, "NSE Stock Market Prediction Using Deep-Learning Models," in *Procedia Computer Science*, 2018, vol. 132, pp. 1351–1362.

[23] Y. S. Park and S. Lek, *Artificial Neural Networks: Multilayer Perceptron for Ecological Modeling*, vol. 28. Elsevier, 2016.

[24] T. Marwala, "Multi-layer Perceptron," *Handb. Mach. Learn.*, no. 2001, pp. 23–42, 2018.

[25] J. Gamboa, "Deep Learning for Time-Series Analysis," *arXiv*, 2017.

[26] P. Gao, R. Zhang, and X. Yang, "The application of stock index price prediction with neural network," *Math. Comput. Appl.*, vol. 25, no. 3, 2020.

[27] W. Lu, J. Li, Y. Li, A. Sun, and J. Wang, "A CNN-LSTM-based model to forecast stock prices," *Complexity*, vol. 2020, 2020.

[28] J. M. Nazzal, I. M. El-emary, S. a Najim, A. Ahliyya, P. O. Box, and K. S. Arabia, "Multilayer Perceptron Neural Network ( MLPs ) For Analyzing the Properties of Jordan Oil Shale," *World Appl. Sci. J.*, vol. 5, no. 5, pp. 546–552, 2008.

[29] G. Van Houdt, C. Mosquera, and G. Nápoles, "A review on the long short-term memory model," *Artif. Intell. Rev.*, vol. 53, no. 8, pp. 5929–5955, Dec. 2020.

[30] Ferdiansyah, S. H. Othman, R. Zahilah Raja Md Radzi, D. Stiawan, Y. Sazaki, and U. Ependi, "A LSTM-Method for Bitcoin Price Prediction: A Case Study Yahoo Finance Stock Market," *ICECOS 2019 - 3rd Int. Conf. Electr. Eng. Comput. Sci. Proceeding*, no. March 2020, pp. 206–210, 2019.

[31] M. Lechner and R. Hasani, "Learning Long-Term Dependencies in Irregularly-Sampled Time Series," *arXiv*, 2020.

[32] H. Wang, Z. Yang, Q. Yu, T. Hong, and X. Lin, "Online reliability time series prediction via convolutional neural network and long short term memory for service-oriented systems," *Knowledge-Based Syst.*, vol. 159, pp. 132–147, 2018.

[33] J. Lu, Q. Zhang, Z. Yang, and M. Tu, "A hybrid model based on convolutional neural network and long short-term memory for short-term load forecasting," *IEEE Power Energy Soc. Gen. Meet.*, vol. 2019-August, 2019.

[34] A. K. Jain, C. Grumber, P. Gelhausen, I. Häring, and A. Stolz, "A Toy Model Study for Long-Term Terror Event Time Series Prediction with CNN," *Eur. J. Secur. Res.*, vol. 5, no. 2, pp. 289–309, 2020.

[35] S. S. Baek, J. Pyo, and J. A. Chun, "Prediction of water level and water quality using a cnn-lstm combined deep learning approach," *Water (Switzerland)*, vol. 12, no. 12, 2020.

[36] S. Selvin, R. Vinayakumar, E. A. Gopalakrishnan, V. K. Menon, and K. P. Soman, "Stock price prediction using LSTM, RNN and CNN-sliding window model," in *2017 International Conference on Advances in Computing, Communications and Informatics, ICACCI 2017*, 2017, vol. 2017-Janua, pp. 1643–1647.

[37] C. Yang, J. Zhai, G. Tao, and P. Haajek, "Deep Learning for Price Movement Prediction Using Convolutional Neural Network and Long Short-Term Memory," *Math. Probl. Eng.*, vol. 2020, 2020.

[38] S. Mehtab and J. Sen, "Stock Price Prediction Using CNN and LSTM-Based Deep Learning Models," *2020 Int. Conf. Decis. Aid Sci. Appl. DASA 2020*, pp. 447–453, 2020.

[39] J. M. T. Wu, Z. Li, N. Herencsar, B. Vo, and J. C. W. Lin, "A graph-based CNN-LSTM stock price prediction algorithm with leading indicators," *Multimed. Syst.*, no. Special Issue Paper, 2021.

[40] A. J. Dautel, W. K. Härdle, S. Lessmann, and H.-V. Seow, "Forex exchange rate forecasting using deep recurrent neural networks," *Digit. Financ.*, vol. 2, no. 1, pp. 69–96, 2020.

[41] A. S. Lundervold and A. Lundervold, "An overview of deep learning in medical imaging focusing on MRI," *Z. Med. Phys.*, vol. 29, no. 2, pp. 102–127, May 2019.

[42] E. Lewinson, "Python for Finance Cookbook," in *Over 50 recipes for applying modern Python libraries to financial data analysis*, 1st ed., Packt Publishing, 2020, p. 434.

[43] K. Wang, K. Li, L. Zhou, Y. Hu, and Z. Cheng, "Multiple convolutional neural networks for multivariate time series prediction," *Neurocomputing*, vol. 360, pp. 107–119, 2019.

[44] E. Hoseinzade and S. Haratizadeh, "CNNpred: CNN-based stock market prediction using a diverse set of variables," *Expert Syst. Appl.*, vol. 129, pp. 273–285, 2019.

[45] L. Ni, Y. Li, X. Wang, J. Zhang, J. Yu, and C. Qi, "Forecasting of Forex Time Series Data Based on Deep Learning," *Procedia Comput. Sci.*, vol. 147, pp. 647–652, 2019.

[46] I. Halimi, G. I. Marthasari, and Y. Azhar, "Prediksi Harga Emas Menggunakan Univariate Convolutional Neural Network," *J. Repos.*, vol. 1, no. 2, p. 105, 2019.

[47] A. Vidal and W. Kristjanpoller, "Gold volatility prediction using a CNN-LSTM approach," *Expert Syst. Appl.*, vol. 157, 2020.

[48] I. E. Livieris, E. Pintelas, and P. Pintelas, "A CNN–LSTM model for gold price time-series forecasting," *Neural Comput. Appl.*, vol. 32, no. 23, pp. 17351–17360, 2020.

[49] R. Yamashita, M. Nishio, R. K. G. Do, and K. Togashi, "Convolutional neural networks: an overview and application in radiology," *Insights Imaging*, vol. 9, no. 4, pp. 611–629, Aug. 2018.

[50] S. Singhal, H. Kumar, and V. Passricha, "Prediction of Heart disease using DNN," *Am. Interantional J. Res. Sci. Technol. Eng. Math.*, no. November, pp. 257–261, 2018.

[51] G. T. Taye, H. J. Hwang, and K. M. Lim, "Application of a convolutional neural network for predicting the occurrence of ventricular tachyarrhythmia using heart rate variability features," *Sci. Rep.*, vol. 10, no. 1, pp. 1–7, 2020.

[52] M. Afrasiabi, H. khotanlou, and M. Mansoorizadeh, "DTW-CNN: time series-based human interaction prediction in videos using CNN-extracted features," *Vis. Comput.*, vol. 36, no. 6, pp. 1127–1139, 2020.

[53] P. Liu, J. Liu, and K. Wu, "CNN-FCM: System modeling promotes stability of deep learning in time series prediction," *Knowledge-Based Syst.*, vol. 203, p. 106081, 2020.

[54] Z. Zhang, Y. Dong, and Y. Yuan, "Temperature Forecasting via Convolutional Recurrent Neural Networks Based on Time-Series Data," *Complexity*, vol. 2020, 2020.

[55] A. G. Salman, B. Kanigoro, and Y. Heryadi, "Weather Forecasting using Deep Learning Techniques," *ICACSIS*, pp. 281–285, 2015.

[56] T. T. Kieu Tran, T. Lee, J. Y. Shin, J. S. Kim, and M. Kamruzzaman, "Deep learning-based maximum temperature forecasting assisted with meta-learning for hyperparameter optimization," *Atmosphere (Basel).*, vol. 11, no. 5, pp. 1–21, 2020.

[57] Z. Alameer, M. A. Elaziz, A. A. Ewees, H. Ye, and Z. Jianhua, "Forecasting gold price fluctuations using improved multilayer perceptron neural network and whale optimization algorithm," *Resour. Policy*, vol. 61, no. September 2018, pp. 250–260, 2019.